%% file: egpaper_for_review.tex
\documentclass[10pt,twocolumn,letterpaper]{article}

\usepackage{cvpr}
\usepackage{times}
\usepackage{epsfig}
\usepackage{graphicx}
\usepackage{amsmath}
\usepackage{amssymb}
\usepackage{bm}
\usepackage{subcaption}

\usepackage[pagebackref=true,breaklinks=true,letterpaper=true,colorlinks,bookmarks=false]{hyperref}

\cvprfinalcopy 

\def\cvprPaperID{****} 

\begin{document}

\title{GAN Based Top-Down View Synthesis in Reinforcement Learning Environments}

\author{Usama Younus\\
University of Maryland, College Park\\
{\tt\small uyounus@umd.edu}
\and
Vinoj Jayasundara\\
University of Maryland, College Park\\
{\tt\small vinoj@umd.edu}
\and
Shivam Mishra\\
University of Maryland, College Park\\
{\tt\small smishra8@umd.edu}
\and
Suleyman Aslan\\
University of Maryland, College Park\\
{\tt\small aslan@umd.edu}
}

\maketitle{}

\section{Introduction}
Human actions are based on the mental perception of the environment. Even when all the aspects of an environment are not visible, humans have an internal mental model that can generalize the partially visible scenes to fully constructed and connected views. This internal mental model uses learned abstract representations of spatial and temporal aspects of the environments encountered in the past.

Artificial agents in the reinforcement learning environments also benefit by learning a representation of the environment from experience. It provides the agent with the viewpoints that are not directly visible to it, helping it make better policy decisions. It can also be used to predict the future states of the environment.

This project explores learning the top-down view of an RL-environment based on the artificial agent's first-person view observations with a generative adversarial network(GAN). The top-down view is useful as it provides a complete overview of the environment by building a map for the entire environment. It provides information about the objects' dimensions and shapes along with their relative positions with one another. Initially, when only a partial observation of the environment is visible to the agent, only a partial top-down view is generated. As the agent explores the environment through a set of actions, the generated top-down view becomes complete. This generated top-down view can assist the agent in deducing better policy decisions. The focus of the project is to learn the top-down view of an RL-environment. It doesn't deal with any Reinforcement Learning task.


\section{Related Work}

The project is based on \cite{DBLP:journals/corr/abs-1803-10122}, which built generative neural network models of popular reinforcement learning environments, and utilized it to learn spatial and temporal representations of the environment in an unsupervised manner for training a simple, compact policy for the given RL task. Other related work includes \cite{Kim2020_GameGan} and \cite{Eslami1204}.

\cite{Eslami1204} introduced Generative Query Network, which takes images for an environment from different viewpoints as input, constructs an internal representation of the environment, and then predicts the environment's appearance from previously unobserved viewpoints. Similarly, \cite{Kim2020_GameGan} learned to visually imitate a game environment by ingesting agents' screenplay and keyboard actions. In this work, they also design a memory module that builds an internal map of the environment to better capture long-term consistency.

\section{Methodology}
\input{sections/methodology}

\input{sections/proposed_method}

\section{Experiments and Results}
\input{sections/results}

\section{Conclusion}

Based on our baseline results, our team attributed the limited scale-ability and generalization to the limited amount of training data that we have for a given set of environment, as well as the lack of proper encoding capability for the temporal domain. For immediate resolution, we proposed two set of variations in our model for tapping into the temporal domain understanding. These included the use 3-dimensional convolutions for capturing the depth and distance relationships, as well as the the use of 2-dimensional followed by 1-dimensional convolutions to capture the spatial and temporal information respectively. For resolving the low training data problem, we suggested the usage of encoders that work well with less amounts of data, while imposing the spatial constraints such as the capsule encoder.

As demonstrated in the experimentation segment, two of these approaches (3D convlutions and Capsule Encoders) were able to make a significant progress towards resolving the  problem of having a complete representation of unseen spatio-temporal scenes in an RL-environment. We were able to show that these encoder variations provide a more representative and realistic top-down view than others; and these views provide an unseen aspect to learn from in addition to the frontal views seen by the agent. These top-down views can be used in a more learned and informed decision policy process by an agent in any environment, and can lead to better decision making. 

A few things that we believe that can be incorporated into our approach is try some additional architectural variations. For our current setup, we were only exploring a random policy by the agent, which in-turn presented a harder training set to learn from. We would like to explore a more informed decision policy for the environment. We would also like to use LSTM in our overall model architecture to retain the environment history and to maintain the overall consistency of the unseen top-down views. Though our model is able to retain this information right now to some extent, but we believe that given a large amount of spatio-temporal information, some inconsistencies might arise from the only generative approach that we utilize.


{\small
\bibliographystyle{ieee}
\bibliography{egbib}
}

\end{document}

%% file: sections/methodology.tex

%% file: sections/proposed_method.tex
We propose a top-down image generation model from the agent's state observations using a GAN based architecture. Our method is different from other methods such as Generative Query Network~\cite{Eslami1204} and World Models~\cite{DBLP:journals/corr/abs-1803-10122} since these methods try to generate images similar to the agent's observations whereas our method can synthesize a new view of the environment. As we have a novel dataset and a novel problem, we first propose a method that we call ``baseline'' and then we also propose improvements to the model to provide comparisons.

We design the neural network architecture to accept a set of subsequent observations of an episode. Let $s_i$ be the set of observations at time step $i$, then $s_i = \{ o_{i}, o_{i-1}, \ldots, o_{i-n}\}$ where $o_i$ is the observation at time $t$. $s_i$ is a padded array, i.e., when the agent has only a single observations at the start of an episode, the set $s_i$ is filled with ``blank'' images as the previous observations. In order to obtain the observations, we simply rotate the camera in the environment around $y$-axis. In our experiments, we set $n=20$ and observations are $64 \times 64$ RGB images. As a result, the input to the neural network is volumetric data with a shape of $(21, 64, 64, 3)$.

Standard GAN architectures generate images from noise, however, in our case we have an input data and at the same time, our problem is different from the image-to-image translation problems so methods such as pix2pix~\cite{isola2017image} and CycleGAN~\cite{zhu2017unpaired} are not suitable. Therefore we define a baseline method from scratch. First, we propose to have an Encoder module before the Generator in order to learn representations from the sets of observations. Our baseline encoder is a 9 layer Convolutional Neural Network (CNN). The Encoder consists of 2D convolutions and in order to feed the volumetric data to the Encoder, we simply stack the channels of all frames, resulting in an ``image'' with $63$ channels. This Encoder outputs $4096$-dimensional features representing the state of the environment based on the observations. The Generator takes these $4096$-dimensional features and generates an RGB image corresponding to the top-down view of the environment for the given time step. We also make modifications to the Discriminator because the ground-truth top-down views are dependent on the state of the environment. As we already have a representation of the state of the environment with a set of observations, we use this set in the Discriminator as well. We apply the same channel stacking to obtain an ``image'' with $63$ channels and we also concatenate this image with the ground-truth RGB image before the first layer of the Discriminator.

Our GAN incorporates several recently proposed improvements. First, we use convolutional GANs with architectural constraints as in~\cite{radford2015unsupervised} to improve the stability of GAN training and the perceptual quality of the generated samples. Arjovsky et al.~\cite{arjovsky2017wasserstein} propose a class of GANs called Wasserstein GAN (WGAN) which defines the objective as:
$$
\min_G\max_{D \in \mathcal{D}} \mathbb{E}_{\bm{x} \sim \mathbb{P}_r}[D(\bm{x})] - \mathbb{E}_{\bm{\Tilde{x}} \sim \mathbb{P}_g}[D(\bm{\Tilde{x}})],
$$
where $\mathcal{D}$ is the set of 1-Lipschitz functions. WGANs alleviate the mode collapse problem, improve the balance in training of the generator and the discriminator, and they are less dependent on the architectural constraints. In~\cite{arjovsky2017wasserstein}, authors enforce the Lipschitz constraint by clipping the weights of the discriminator after each gradient update. We use the same objective as WGAN however we use a different method to enforce the Lipschitz constraint. In WGAN-GP~\cite{gulrajani2017improved}, which is another form of WGAN, the Lipschitz constraint is enforced by the gradient penalty instead of weight clipping. This provides stronger modeling performance and improved stability and the objective becomes:
\begin{multline*}
\min_G\max_{D} \mathbb{E}_{\bm{x} \sim \mathbb{P}_r}[D(\bm{x})] - \mathbb{E}_{\bm{\Tilde{x}} \sim \mathbb{P}_g}[D(\bm{\Tilde{x}})] \\ - \lambda \mathbb{E}_{\bm{\hat{x}} \sim \mathbb{P}_{\bm{\hat{x}}}}[(||\nabla_{\bm{\hat{x}}}D(\bm{\hat{x}})||_2-1)^2],
\end{multline*}
where $\lambda$ is the penalty coefficient and $\mathbb{P}_{\bm{\hat{x}}}$ is defined as uniformly interpolating between pairs of samples from $\mathbb{P}_r$ and $\mathbb{P}_g$. We use this objective in our methods.

We also incorporate the training methodology for GANs introduced in Progressive Growing of GANs (PGAN)~\cite{karras2017progressive}. In this training methodology, the generator and the discriminator start from a low resolution and grow progressively by adding new layers. \cite{karras2017progressive} inserts mini-batch standard deviation feature map in the discriminator to improve variation and uses equalized learning rate as a normalization in the generator and the discriminator as well. Additionally, PGAN applies pixel-wise feature vector normalization after each convolutional layer in the generator. Our GAN includes all of these improvements. We use Adam optimizer~\cite{kingma2014adam} for the Generator (including the Encoder) and the Discriminator. For the training, we start at $4 \times 4$ size images and we increase the scale by doubling the size at every 25,000 iterations up to the original image size $64 \times 64$. As explained in~\cite{karras2017progressive}, we linearly interpolate the output weight of the new layers $\alpha$ from 0 to 1 at 12,500 iterations when we increase the scale. We publish our method through a GitHub repository\footnote{\url{https://github.com/suleymanaslan/top-down-view-gan}}.

\begin{figure}[h]
\centering
\includegraphics[width=\columnwidth]{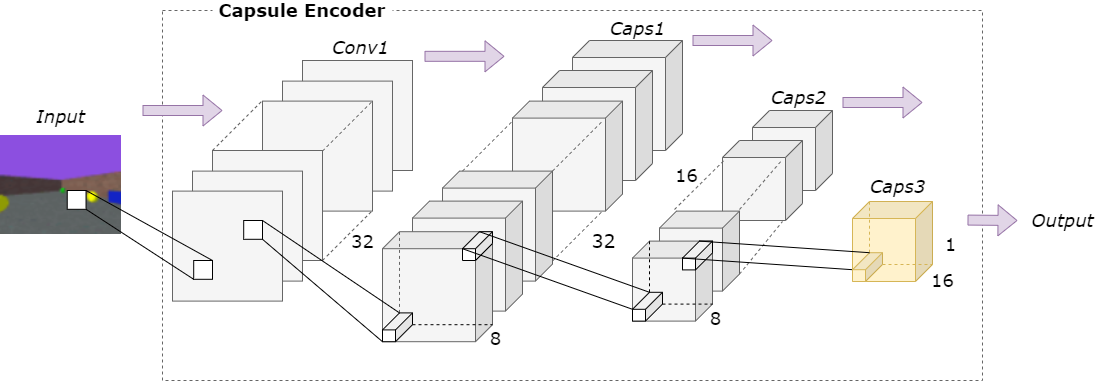} 
\caption{The capsule encoder architecture. The input is fed to the capsule encoder, which passes a concise representation of the input to the subsequent parts of the system.}
\label{fig:capsule_encoder}
\end{figure}

\begin{figure*}[h]
    \centering
    \subcaptionbox{Observations.}{{\includegraphics[width=0.5\linewidth]{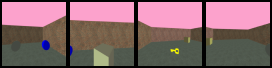}}{\includegraphics[width=0.5\linewidth]{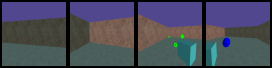}}}
    \subcaptionbox{Generated images.}{{\includegraphics[width=0.5\linewidth]{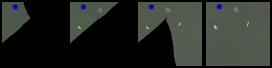}}{\includegraphics[width=0.5\linewidth]{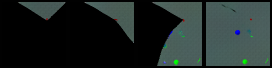}}}
    \caption{Sample observations of the agent in the training environments and the model predictions corresponding to the top-down views at the same time step.}
    \label{fig:sample_images_training}
\end{figure*}

Based on our initial set of experimentation, we propose some  improvements  in  the  model  architecture  due  to  the limited learning capability of our baseline approach. Our baseline was not able to generalize well on the test dataset. Additionally, it wasn’t able to capture the spatial distance of the environment in its entirety, i.e. it was unsuccessful in complete articulation of the inter-object distances. Henceforth, we propose to use some variants of the encoder. The baseline encoder does not take the temporal information into account. Therefore we focus on learning better representations with an encoder that utilizes the spatio-temporal information. For this purpose, we first use an encoder with 3D convolutions. As the depth (time) dimension is smaller than the width and height, the size of the convolving kernels that we use are smaller in the temporal dimension. Additionally, unlike the padding added to the image plane, we do not add padding for the depth dimension. The learned features are $4096$-dimensional, same as before. Therefore we do not to make any changes to the generator. In another variant, we separate the processing of spatial and temporal information and we use an encoder that consists of 2d convolutions to extract spatial features followed by 1d convolutions to extract temporal features. In this method, the spatial CNN layers are applied per-frame basis and this CNN outputs features for each frame, resulting in a $21 \times 4096$ dimensional feature map. Then, the temporal CNN takes this feature map and processes the features at each time step by applying 1D convolutions. As the output size is same as before, we do not change the generator for this method as well. 

Evidently, it is advantageous for the task at hand, if the encoder itself supplements learning in the low-data regime. CNNs are generally translation invariant, and not invariant to other transforms such as the orientation changes. Hence, they require a lot of training data with ample variations to learn to handle such transforms, resulting in reduced generalization capabilities. In contrast, CapsNets are equivariant, where lower level capsules exhibit place-coded equivariance and higher level capsules exhibit rate-coded
equivariance \cite{sabour2017dynamic}. For instance, CNNs learn orientation invariance by training on a large number augmented images, whereas CapsNets learn to encode orientation in their instantiation parameters without observing many such augmentations. Subsequently, CapsNets are able to successfully encode orientation, even for an image outside the
training domain. Hence, we argue that CapsNets show better generalization for the task at hand, compared to conventional CNN models. Furthermore, CapsNets learn the part-whole spatial relationships between the entities present in the input, enforcing the preservation of spatial constraints between entities present in the input. Hence, we argue that a capsule encoder can significantly enhance the performance of our system, since it addresses the low-data problem as well as issue of distorted spatial relationships present in the baseline method. 

As illustrated in Fig. ~\ref{fig:capsule_encoder}, the proposed capsule encoder is adopted from \cite{sabour2017dynamic}, and consists of a convolutional layer having kernels of size 3 $\times$ 3 and a Leaky ReLU activation, followed by three convolutional capsule layers, with the number of channels of the last layer set to $1$. The \textit{Caps1} layer consists of 32 channels of 8-dimensional capsules, which will be dynamically routed to each of the 16 channels of 8-dimensional capsules in the \textit{Caps2} layer. We adopt the dynamic routing algorithm proposed in \cite{sabour2017dynamic} for routing. The decision to keep the output of the capsule encoder to one channel in the \textit{Caps3} layer stems from the capsule representational assumption. 
Hence, at each location in the input image, there is at most one instance of the type of entity that a capsule represents. 
This allows us to have a single capsule channel in our representation.

\begin{figure*}[h]
    \centering
    \subcaptionbox{Observations.}{{\includegraphics[width=0.5\linewidth]{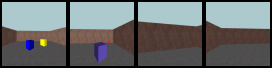}}{\includegraphics[width=0.5\linewidth]{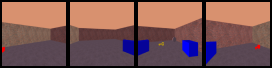}}}
    \subcaptionbox{Generated images.}{{\includegraphics[width=0.5\linewidth]{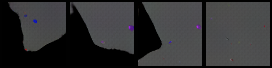}}{\includegraphics[width=0.5\linewidth]{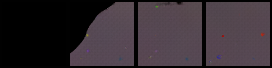}}}
    \caption{Sample observations of the agent in the testing environments and the model predictions corresponding to the top-down views at the same time step. Compared to the training results, the method is not able to generalize well.}
    \label{fig:sample_images_testing}
    \vspace{2mm}
\end{figure*}

\begin{table*}
\caption{The comparison of average PSNR and SSIM values for the training and testing sets.}
\label{tbl:comparison}
\begin{center}
\begin{tabular}{|l|c|c|c|c|}
\hline
\textbf{Method} & \multicolumn{2}{c|}{\textbf{Average PSNR}} & \multicolumn{2}{c|}{\textbf{Average SSIM}} \\
\hline\hline
& Training & Testing & Training & Testing\\
\cline{2-5}
Baseline & 30.0 $\pm$ 1.8 & 16.9 $\pm$ 2.0 & 0.9251 $\pm$ 0.0132 & 0.5584 $\pm$ 0.1195\\
Ours (with 2D+1D encoder) & - & - & - & -\\
Ours (with 3D encoder) & 31.4 $\pm$ 1.7 & 18.4 $\pm$ 2.1&  \textbf{0.9481 $\pm$ 0.0142} & 0.6617 $\pm$ 0.1003\\
Ours (with capsule encoder) & \textbf{31.8 $\pm$ 2.2} & \textbf{20.6 $\pm$ 2.3} & 0.9376 $\pm$ 0.0247 & \textbf{0.7123 $\pm$ 0.1850}\\
\hline
\end{tabular}
\end{center}
\end{table*}

%% file: sections/results.tex
In this section, we compare the results generated by our proposed baseline, as well as the additional improvements in our baseline for a more cohesive generative learning of the agent's environment.

We first evaluate the proposed method qualitatively. Based on our experiments, we observe that the baseline method can learn to generate the top-down views from the training data. The outputs of the model show that the proposed method can generate the image with correct scene lighting, object locations, shapes, and colors. The method can also distinguish the observed parts of the scene from the unseen parts during the earlier time steps and it remembers the previously seen objects even if they are no longer visible at later time steps. Sample images for the training environments results are given in Figure~\ref{fig:sample_images_training}. However, when we evaluate the method in the unseen (testing) environments, we observe that it does not perform well, therefore the model is unable to generalize. Results for the testing environments are given in Figure~\ref{fig:sample_images_testing}, it can be seen that the test performance differs greatly from the training performance. These initial experiments demonstrate that this model has an overfitting problem. Therefore we compare the other proposed methods with the baseline in terms of testing performance. For this purpose, we provide quantitative evaluation. However, during the training of method with 2D+1D encoder, we observe that this model is not able to learn from the training data. Because of this reason, we exclude this model from the comparisons as this model does not work for this problem. We believe that, the 1D CNN is not able to keep the useful learned spatial features and the output of 1D convolutions do not produce accurate representations for the state of the environment for the generator. 

The metrics we have used for the quantitative evaluation of our results are: PSNR \& SSIM. The results are given in Table~\ref{tbl:comparison}. Since our approach intends to create an unseen view of a spatially consistent environment, we use PSNR (peak signal-to-noise ratio) as an evaluation metric to access the quality of the top-down views generated by our model. This metric evaluates this by computing the power ratio of the ground-truth image with the generated image; thus, providing a decent reconstruction evaluation metric. Another metric we have used for the quantitative evaluation of our model's performance is SSIM (structural similarity index). This metric enables us to evaluate the similarity between the generated top-down view and the ground-truth, along with an evaluation of the quality degradation as a perceived change in structural information of the image. SSIM utilizes a series of Gaussian Window operations on luminescence, contrast and structure to create a quality map of the image.

As evident from Table \ref{tbl:comparison}, the 3D encoder-based method outperformed the baseline, benefiting from the additional temporal information as hypothesized. Furthermore, the capsule encoder-based method significantly outperformed both the baseline and the 3D encoder-based method, in terms of both PSNR and SSIM. As hypothesized, the capsule encoder has performed the best out of the compared models, due to its inherent capabilities to work with low data, and the spacial constraints imposed by learning the part-whole relationships between entities. It is interesting to note that the training average SSIM for the 3D encoder-based model is better than that of the capsule encoder model, yet, significantly lower in the case of the testing average SSIM. This suggests that given similar training settings (including the number of training iterations), the 3D encoder model probably overfits to the training data more than the capsule encoder model, which better generalizes to the testing set.